# Building Arabic NLP from the Ground Up: Twenty Years of Lessons, Failures, and Open Problems

**Wajdi Zaghouani**
Communication Program
Northwestern University in Qatar
Doha, Qatar
wajdi.zaghouani@northwestern.edu

## Abstract

This paper reflects on twenty years of building NLP resources and research infrastructure for Arabic, a language spoken by hundreds of millions yet historically underserved relative to languages such as English or Chinese. The first decade focused on foundational linguistic infrastructure; the second shifted toward computational social science, social media analysis, and socially oriented applications. Rather than cataloguing outputs, the paper examines what the experience of building them revealed. Three counterintuitive lessons emerge: building datasets is as much a social process as a technical one; communities formed around shared tasks often matter more than the tasks themselves; and moving from language resources to computational social science exposes challenges that traditional NLP training does not address. We discuss three failures: a depression detection corpus that never reached clinical practice, a period of spreading across too many shared tasks without sufficient depth, and a long-standing assumption that Modern Standard Arabic infrastructure would transfer cleanly to dialectal tasks. These experiences suggest that the hardest problems in developing NLP for underserved communities are not linguistic but social, institutional, and epistemic, and require competencies the field rarely teaches.

## 1 Why Write This Paper

Most papers in NLP report successes. A new dataset achieves higher coverage than its predecessor. A new model beats the previous state of the art. A shared task attracts more participants than the last edition. This is how the field communicates progress, and it is broadly fine. But it produces a literature with a systematic blind spot: the accumulated wisdom about what does not work, what surprised the people who built things, and what they would do differently, rarely gets written down in a form anyone else can learn from.

When something fails in NLP, the failure is usually absorbed quietly. The dataset sits on a server and is not cited. The shared task runs once and is not repeated. The deployment pilot ends without a follow-up paper. The PhD student who ran the failed experiment writes up a different result instead. None of this is dishonest, exactly, but the cumulative effect is a literature that tends to overrepresent successful results relative to failed experiments, a pattern well documented in meta-science more broadly (Fanelli, 2012). For senior researchers reflecting on a body of work, there is also a subtler pressure: the narrative of a research program tends toward coherence and inevitability in retrospect, because that is how people tell stories about themselves. Fighting that pressure is one reason this paper exists.

This paper attempts to fill some of that gap for one particular research program: twenty years of building Arabic NLP, spanning roughly a first decade of foundational linguistic infrastructure and, beginning in the mid-2010s, an expanding focus on social media analysis, computational social science, and socially oriented applications in and around the Arab world. The program is long enough now to look back on honestly. Some things worked better than expected. Some things failed. Some failures were obvious in retrospect. Some successes turned out to matter less than anticipated. And a few lessons emerged that seem genuinely transferable to any researcher trying to build NLP for an underserved language or community.

The paper is organized around three questions the Big Picture workshop explicitly invites: What was the larger vision? What worked and what did not? And what comes next, not just for this program but for the set of problems it represents?

One framing note before proceeding: nothing in this paper is specific to Arabic in a way that makes it irrelevant to other contexts. The challenges of building language infrastructure for a morphologi-cally complex, dialectally diverse, politically sensitive, and historically under-resourced language generalize. Researchers working on low-resource African languages, South Asian languages, Indigenous languages of the Americas, or regional languages of Europe will recognize most of what follows. That generalizability is part of the point.

The paper situates itself within a body of work interrogating how NLP infrastructure gets built and for whom. Proposals like datasheets for datasets (Gebru et al., 2021) and data statements for NLP (Bender and Friedman, 2018) advocate for transparency in resource documentation. Critiques of benchmark culture (Bowman and Dahl, 2021; Ethayarajh and Jurafsky, 2020) have questioned whether leaderboard-driven evaluation advances understanding or merely advances numbers. Participatory approaches to ML (Birhane et al., 2022) argue that communities whose language is being studied should shape what is built and how. This paper operates in the same register: its claims are grounded in practice, and it is offered as a position piece that others can test against their own programs.

## 2 The Vision: What We Thought We Were Building

The research program began, like many in NLP, with a resource gap. In the mid-2000s, Arabic was a language spoken by hundreds of millions of people yet had no publicly available syntactically annotated corpus comparable to those for English, very limited large-scale error-corrected text, almost no social media datasets, and few evaluation benchmarks. The initial vision was straightforward: build the infrastructure, and the research will follow.

That vision was correct in the narrow sense. Infrastructure does enable research. But it was incomplete in a way that took years to fully understand: infrastructure does not automatically enable the *right* research, by the *right* people, for the *right* purposes. It enables whatever research the people who have access to it choose to do, which is not always what the people who built it imagined. This gap between what a resource is built for and what it is actually used for is probably universal in dataset work, but it is especially consequential in low-resource language NLP, where each resource represents a significant fraction of the total infrastructure and therefore shapes downstream research directions more heavily.

Roughly the first decade of the program, from 2004 to 2014, was spent primarily on foundational resources: the Arabic Treebank series (Maamouri et al., 2010; Zaghouani, 2010), the Arabic PropBank (Zaghouani et al., 2010; Palmer et al., 2008), named entity recognition systems (Zaghouani, 2012; Zaghouani et al., 2010), error correction corpora and shared tasks (Zaghouani et al., 2014; Mohit et al., 2014; Rozovskaya et al., 2015), morphological resources (Zaghouani et al., 2016), and dialect corpora (Bouamor et al., 2018). These were necessary. They also consumed enormous time, and the vision of what they were *for* evolved considerably during their construction.

Beginning in the mid-2010s, the program expanded toward using those resources to study real social phenomena: who speaks Arabic online and how (Zaghouani and Charfi, 2018; Rangel et al., 2020); how hate speech and misinformation spread in Arabic social media (Charfi et al., 2024; Zaghouani et al., 2024); how mental health signals appear in Arabic text (Zaghouani, 2018; Zaghouani and Biswas, 2025); how political discourse is framed in Arabic news and social media (Shurafa et al., 2020; Shestakov and Zaghouani, 2024); and what digital citizenship means for Arabic-speaking communities (Al Heraki and Zaghouani, 2025; Zaghouani et al., 2026).

The transition from resource building to social analysis felt natural from inside the program. It was not a pivot; it was a maturation. But looking back, it is clear that the two phases required fundamentally different competencies, and assuming the first would automatically prepare you for the second was a mistake that cost time and produced some work that was thinner than it should have been. There is also a vision-level failure worth naming explicitly: as the program moved into socially oriented work in its second decade, we started with the implicit assumption that building better NLP tools was sufficient, and that social scientists and policymakers would eventually discover and use them. This assumption is widespread in the field, and it is largely false. Policymakers rarely engage directly with NLP research publications, and social scientists are not waiting for a better hate speech classifier to appear in an ACL proceedings volume. The gap between producing a technical result and having that result affect anything outside academia is enormous, and bridging it requires forms of engagement that NLP researchers are not trained for and that academic incentive structures do not re-ward. Recognizing this earlier would have changed how several projects were designed.

## 3 What Worked: Three Counterintuitive Lessons

### 3.1 Datasets Are Social Infrastructure, Not Just Technical Artifacts

The first lesson sounds obvious but took years to fully internalize: the value of a dataset is not in the data. It is in the community that forms around it.

The Qatar Arabic Language Bank (QALB) (Zaghouani et al., 2014) was designed as an error correction corpus. Its technical contribution was well defined: a large collection of manually corrected Arabic texts with detailed annotation guidelines (Zaghouani et al., 2015). By any standard metric of dataset quality, it was good. But what made it genuinely impactful was not the corpus itself. It was the shared tasks we organized around it at EMNLP 2014 (Mohit et al., 2014) and ACL 2015 (Rozovskaya et al., 2015), which brought together teams who would not otherwise have had reason to work on Arabic error correction. The task created a reason to engage with the data, a competitive structure that motivated effort, and a venue where results could be compared and discussed. Remove the shared task and QALB is a dataset. Add the shared task and QALB becomes a coordination mechanism for a research community.

The pattern repeated with the AraP-Tweet corpus (Zaghouani and Charfi, 2018) and the author profiling shared task at FIRE 2019 (Rangel et al., 2019), and again with ADHAR (Charfi et al., 2024), the MAHED multimodal shared task (Zaghouani et al., 2025), and ImageEval 2025 (Bashiti et al., 2025). In every case, the dataset alone attracted limited engagement. The shared task built around the dataset created a network. Some of those networks persist long after the task itself has ended: collaborations that formed during a shared task evaluation cycle have continued for years, producing work that has nothing to do with the original task.

This has a practical implication that the NLP literature rarely discusses: if you are building a dataset for an underserved language and you want it to have impact beyond your own group, the data release is not the end of the work. It is the beginning of a community-building project that requires sustained organizational effort, outreach, and maintenance. Researchers who treat dataset release as the finish line often find that their corpora are cited but not actually used in downstream work by people outside their circle. Citation is not impact. Actual use by people who were not involved in building it is closer to impact.

The Arabic Natural Language Processing Workshop (WANLP) series, which we have co-organized since 2014 (Habash et al., 2017; Zitouni et al., 2020; Habash et al., 2021; Bouamor et al., 2022), illustrated this at workshop scale. The technical papers at WANLP are individually unremarkable by top-conference standards. What WANLP built over a decade was a community: a set of people who know each other, share datasets informally, collaborate across institutions, and maintain shared standards about what counts as good work in Arabic NLP. That community is more durable than any individual resource. When a new student arrives wanting to work on Arabic NLP, they do not start from scratch; they start from a community with accumulated norms, shared baselines, and accessible mentorship. Building that takes a decade. It cannot be replicated by a grant or a paper.

### 3.2 The Shared Task Is a Research Instrument, Not Just an Evaluation Exercise

The second lesson follows from the first and is equally underappreciated: shared tasks, run well, are one of the most efficient mechanisms for accelerating research in a new area, precisely because they force clarity on questions that individual researchers can avoid indefinitely.

The field tends to think of shared tasks as evaluation exercises. You build a benchmark, teams submit systems, you measure who performs best. That is a reasonable description of what happens technically. But what actually happens socially is different. A well-designed shared task forces a community to agree, at least provisionally, on what a problem *is*, how it should be measured, and what counts as a valid solution. These are contested questions in any active research area, and the process of contesting them in public, with real systems and real results, advances understanding faster than individual papers arguing theoretical positions.

The CheckThat! Lab collaborations (Nakov et al., 2018, 2022; Hasanain et al., 2024), which ran over multiple years on fact-checking, misinformation, and subjectivity detection, demonstrated this clearly. The technical problem of claim check-worthiness estimation sounds well defined until you try to annotate it. Annotation disagreements in early rounds surfaced genuine conceptual ambiguities: is a claim check-worthy because it is important, because it is verifiable, or because it is likely to be believed by a specific audience? These are not annotation errors. They are different theories of what check-worthiness means. Resolving them required cross-team discussion that would never have happened without the shared task as a forcing function. Several papers produced during CheckThat! participation were less about system performance than about reframing what the task should be, which is exactly the kind of output a maturing research area needs.

The lesson for researchers entering new domains is this: if you are building resources for a problem that has not been studied computationally before, organizing a shared task early, before your own models are mature and before the task definition has stabilized, is probably more valuable than publishing another dataset paper. The task will teach you things about your own dataset that your own analysis would not reveal, because other people bring assumptions you did not know you were making.

There is also a less obvious benefit: the shared task as a mentorship infrastructure. For a research group based in a region that is geographically distant from the major NLP conference hubs, shared task co-organization is one of the most effective ways to bring junior researchers into contact with the broader community. Students who would never have had a reason to interact with senior researchers at CMU, NYU, or the University of Edinburgh found those interactions through shared task evaluation discussions. That is not a trivial benefit.

### 3.3 Moving Into Social Science Requires Unlearning Some NLP Habits

The third lesson is the most uncomfortable to write about, because it touches on the limits of the training most NLP researchers receive and the assumptions that training instills without announcing them.

When the research program shifted from building linguistic infrastructure to studying social phenomena, we carried a set of habits from NLP that turned out to be poorly suited to the new domain. The most consequential of these was treating annotation disagreement as noise to be minimized.

In syntactic annotation, inter-annotator disagreement usually means the guidelines are unclear or the annotator made an error. The fix is better guidelines, more training, and adjudication procedures. This worked reasonably well for the Arabic Treebank and for QALB, both of which deal with phenomena that have defensible correct answers. It worked badly for hate speech, stance detection, and emotion recognition.

When we built the ADHAR hate speech corpus (Charfi et al., 2024) and the multi-label hate speech dataset (Zaghouani et al., 2024), we encountered annotation disagreement rates that were high by NLP standards but turned out to be meaningful rather than erroneous. Two Arabic speakers from different countries, educational backgrounds, and political perspectives often genuinely disagreed about whether a statement was offensive, and about what made it so. The disagreement was not a measurement error. It was the phenomenon. The distribution of opinions about what counts as offensive language in Arabic social media *is* the thing we wanted to understand. Collapsing that distribution into a majority-vote gold label discards the most socially important information in the annotation.

The NLP literature has become more sophisticated about this (Plank, 2022), and work on learning from disagreement, soft labels, and annotator modeling is growing. But the field's dominant evaluation framework, built around single gold labels and metrics that assume ground truth, still treats inter-annotator disagreement as a problem to be solved rather than a signal to be preserved. For any task involving human judgment about social phenomena, this is a systematic limitation that produces benchmarks which look clean but measure something subtly different from what the task was supposed to capture.

Unlearning this habit took longer than it should have. The deeper lesson is that moving from language technology to computational social science is not just a change of application domain. It is a change of epistemic framework. Social science has decades of sophisticated thinking about measurement validity, construct validity, and the relationship between operationalization and theory. NLP researchers entering social science domains typically bring none of that training. We did not bring it either, and the work suffered for it in ways that are visible only in retrospect.

# 4 What Did Not Work: Three Honest Failures

## 4.1 The Gap Between Detection and Deployment

In 2018, we published a large-scale social media corpus for Arabic youth depression detection (Zaghouani, 2018). The motivation was clear and genuine: mental health stigma in Arab societies suppresses help-seeking behavior to a degree that has measurable public health consequences. People who would not speak to a doctor do express distress online, and computational tools that could identify at-risk individuals earlier than clinical referral pathways seemed both feasible and valuable.

The corpus was well constructed. The models trained on it achieved reasonable performance on held-out test data. The paper was published and received citations. To our knowledge, no clinical deployment or policy adoption resulted from the work.

Looking back, the failure was not technical. It was a failure to engage, from the beginning, the clinical, ethical, and regulatory infrastructure that would be required for any actual deployment. Mental health detection from social media text raises serious questions that we were not equipped to answer: What is the consent framework for using someone's social media posts to infer their mental health status? What happens when the system produces a false positive? Who is responsible if an at-risk individual is flagged and no clinical response follows? How does a model trained on Arabic text from one country generalize to another where the cultural expression of distress is different? We did not have clinical collaborators, ethical review partnerships, or deployment relationships in place. We had a dataset and a model, which turned out to be the easy part.

This is not an isolated failure. A significant fraction of NLP-for-social-good work follows the same pattern: a technically interesting problem, a corpus, a classifier, a publication, and then silence. The medical NLP literature is full of systems that achieve impressive performance on benchmark datasets and have never been used by a clinician. The gap between benchmark performance and clinical deployment is well documented (Obermeyer and Emanuel, 2016) but rarely discussed candidly in the papers that report the benchmark results. The lesson has since shaped how subsequent mental health NLP work is framed. The MINDSCAPE-QA proposal currently in development begins with clinical collaborators and ethical review structures before the first annotation decision is made. The corpus design is constrained by what a clinical deployment would actually need, not by what is easiest to annotate. This is slower and harder. It is also the only approach that has any chance of producing something that matters outside a benchmark.

Based on this experience, we propose a minimal governance checklist that any mental health NLP project should satisfy before annotation begins: (a) a signed memorandum of understanding with at least one clinical partner who has agreed to participate in deployment planning; (b) IRB or equivalent ethics board approval covering the data collection protocol and annotator wellbeing; (c) a written data minimization policy specifying what is collected, from whom, for how long it is retained, and under what conditions it is deleted; (d) a false-positive response protocol defining what happens when a system flags a user who does not need intervention; and (e) a cultural adaptation review by mental health professionals familiar with the target community, because distress expression varies substantially across Arabic-speaking societies. None of these prerequisites are technically demanding. All of them require institutional relationships that take time to build. Starting that work at the proposal stage, not the publication stage, is the difference between research that could be deployed and research that cannot. The broader point, which is the most important takeaway from this section, is that NLP researchers working on socially oriented problems should integrate domain experts, ethics review boards, and policy partners at the project design stage, not as downstream consultants once the dataset and model already exist. Current practice typically reverses this order, and the cost of that reversal is research that does not translate into use.

## 4.2 The Breadth-Depth Tradeoff in Shared Task Participation

Between 2023 and 2025, MarsadLab participated in a large number of shared tasks across the ArabicNLP community, including AraGenEval, AraHealthQA, BAREC, NADI, PalmX, MAHED, TAQEEM, and others (Biswas et al., 2025; Bessghaier et al., 2025; Ibrahim et al., 2025; Biswas et al., 2025,?; Zaghouani et al., 2025; Bessghaier et al., 2025). The stated goal was twofold: train students through the discipline of a submission deadline, and establish the group's presence across multiple research fronts.

The honest assessment is that this strategy produced many papers but uneven scientific depth. Several of the shared task submissions were primarily engineering exercises: fine-tuning pre-trained Arabic models with task-specific augmentation rather than contributions to understanding the underlying problems. We were optimizing for participation breadth rather than for insight, and the submissions showed it. A student who fine-tunes AraBERT for eight different tasks in a year learns something about practical NLP engineering. They do not develop the deep engagement with a single problem that produces real understanding.

Some of the most interesting problems we touched during this period deserved more sustained attention than a shared task submission cycle allows. Arabic readability, for instance, is a genuinely underexplored problem with real educational implications. Multimodal propaganda detection in Arabic memes raises hard questions about the relationship between visual and linguistic meaning that a shared task run cannot resolve. We produced submissions on both topics. We did not produce the work those topics deserved.

The practical lesson for research groups building capacity through shared tasks is that participation is useful training for junior researchers but is not a substitute for sustained engagement with a problem. There is an important distinction between using shared tasks to train students and using them as a primary publication strategy. The former is defensible. The latter tends to produce thin work that has high paper counts and low scientific impact. If a shared task consistently reveals something genuinely surprising about how models handle a problem, that surprise is worth following into original research. If the result is always approximately what you expected, the submission may not be worth the cost to the students who ran it.

### 4.3 The Assumption That MSA Resources Transfer to Dialectal Tasks

The early years of the program were spent building resources for Modern Standard Arabic (MSA), the formal register used in news, official communication, and education across the Arab world. MSA is important, heavily used in writing and formal speech, and linguistically well defined. The resources built during the treebank and error correction years remain in active use. But those years also embedded an assumption that took a decade to fully dislodge: that MSA infrastructure would transfer, with some adaptation, to the dialectal Arabic that most people actually use when they communicate informally online or in speech.

It does not transfer nearly as well as expected (Abdul-Mageed et al., 2021).

Arabic dialects are not stylistic variations of MSA in the way that formal and informal registers of many languages relate to each other. They have phonological systems that diverge substantially from MSA, distinct vocabulary sets with massive English and French loanword influence in some varieties, morphological patterns that differ systematically, and discourse conventions that MSA text simply does not represent. A model trained on MSA newswire text, even one fine-tuned on dialectal data, consistently makes errors on Gulf Arabic, Moroccan Darija, or Egyptian colloquial that are not just quantitatively worse but qualitatively different: the model fails on constructions that do not exist in MSA and has no framework for handling them.

We recognized this problem relatively early and contributed to the corrective through the MADAR corpus (Bouamor et al., 2018) and dialect orthography guidelines (Habash et al., 2018). But the field as a whole, including this program, continued for too long to treat dialect adaptation as a minor engineering challenge. The framing was: we have MSA resources, we add some dialectal data, we fine-tune, problem partially solved. The more accurate framing is: dialectal Arabic is a set of genuinely distinct language varieties that each require dedicated resource development from the ground up, and the fact that they share a name and a script with MSA is a source of confusion as much as a foundation to build on. This failure generalizes directly to other language communities. Any language with significant internal variation, including Hindi and Urdu, Mandarin and Cantonese, Brazilian and European Portuguese, and written and spoken varieties of Norwegian, is at risk of producing NLP infrastructure that serves its formal prestige variety while leaving the varieties that most speakers actually use in most of their daily lives severely underserved. The field tends to treat the prestige variety as the language and the other varieties as variants, which is a sociolinguistic assumption embedded in resource design decisions, not a neutral technical choice.

## 5 Wider Reflections: What This Means Beyond Arabic

This section steps back from Arabic specifically and asks what the experience of building this research program suggests about NLP more broadly, including for high-resource language research.

**Compressed infrastructure timelines create compounding debt.** English NLP had decades to build its infrastructure before the era of deep learning. Arabic NLP has tried to compress that timeline dramatically, and the compression has costs that accumulate in ways that are hard to see in the moment. When you build a treebank, then immediately build a social media corpus using the same annotation team, then immediately build a hate speech dataset using the same guidelines philosophy, you inherit the assumptions and limitations of each previous step into the next. MSA-centric assumptions persisted into social media annotation longer than they should have, because the same people were doing both and had internalized those assumptions as defaults. Low-resource language NLP programs should build in deliberate pauses for retrospective audit: structured moments to ask whether the assumptions that made sense at step one still make sense at step four. This is almost never funded and almost never done. But the audit is the work that prevents a decade of compounding debt. In practice, such an audit need not be elaborate. A structured two-day workshop involving the annotation team and at least one external reviewer, convened every two to three years, with an explicit mandate to question whether annotation guidelines, dialect coverage, and task framing still reflect the phenomena the program is trying to study, would be sufficient. The cost is low. The expected benefit, in catching assumptions that have quietly stopped being true, is high.

**The hardest problems are not linguistic, they are social.** The problems that most slowed this research program were not morphological complexity or dialectal diversity, though those were genuine and time consuming. They were social: getting clinical partners to engage with mental health NLP on terms that would support actual deployment; getting platform partners to provide data access that did not evaporate when a terms-of-service update changed API policies overnight; getting annotation workers to complete sensitive tasks about hate speech and trauma without suffering psychological harm; getting policymakers to treat computational findings as evidence worth engaging with rather than as academic noise.

None of these problems appear in the NLP curriculum. None of them appear as explicit criteria in grant review panels, which evaluate technical merit and broader impact but rarely require evidence that a team has the social infrastructure to bridge the gap between those two things. And yet all of them determine, more than the technical quality of the models, whether a body of work has any real-world effect.

The lesson is not that NLP researchers should become clinicians or policymakers. It is that teams working on socially oriented NLP should include people with those competencies from the beginning of a project, not as downstream consultants once the technical work is done. This is an argument for interdisciplinary team composition that the field talks about frequently and practices rarely.

**Community infrastructure outlasts any individual resource.** The Arabic Treebank from 2010 will eventually be superseded by larger, more diverse, neural-era corpora. The specific shared tasks from 2014 and 2015 are no longer running. Individual datasets become obsolete as the phenomena they capture shift or as the models that use them are replaced. What does not become obsolete is a research community that knows how to build things together: shared norms about annotation quality, shared practices about data release, shared venues where disagreements can be aired and partially resolved, and personal relationships that make cross-institutional collaboration something other than a bureaucratic exercise.

The most durable investments in this research program were the community investments: WANLP and its surrounding network, the shared task infrastructure, the training and mentorship relationships with students who are now doing independent research. Individual papers report results; communities build fields. For a language community that is trying to build NLP infrastructure from a deficit position, this means that community-building is not a secondary activity to be done after the real research. It is the primary activity, and the research is what sustains it. This is one place where the conventional wisdom of the field is most misleading: career incentives in NLP reward individual technical contributions, while the work that most reliably matures a research area is the unglam-orous coordination work that no single paper can capture.

**Annotator perspective is data, not noise.** This deserves emphasis because the dominant evaluation culture in NLP actively works against it. When annotators disagree about whether a tweet is hate speech, or whether a social media post expresses depression, the standard procedure is majority-vote adjudication. The resulting gold label is clean, model-friendly, and epistemically misleading. If 40% of Arabic speakers consider a statement offensive and 60% do not, that split is a fact about the social reality of hate speech in Arabic-speaking communities. A gold label of "not hate speech" erases that fact. Systems trained on adjudicated labels also learn to replicate the adjudicator's demographic perspective, regardless of what the guidelines said about neutrality, a problem documented for English (Sap et al., 2019; Davidson et al., 2019) but understudied for Arabic. The pressure to produce clean gold labels moreover creates perverse design incentives: guidelines minimize disagreement rather than capture genuine conceptual boundaries, and tasks are scoped to problems annotators will agree on, which tends to mean problems that are less socially significant. The field is slowly learning this (Plank, 2022), but evaluation frameworks built on single gold labels remain the norm. High inter-annotator agreement is a quality signal only conditionally. For tasks involving social judgment, it may mean the task was scoped too narrowly to matter.

A concrete corrective: release per-annotator labels alongside aggregated labels, include annotator demographic summaries in datasheet documentation (Gebru et al., 2021), and report model performance under majority-vote, soft-label, and per-annotator aggregation schemes so that sensitivity to label aggregation is visible rather than assumed away. This is a documentation norm, not a research burden.

## 6 What Comes Next

**From detecting harm to understanding flourishing.** The bulk of this program's social media work has been oriented toward detection: hate speech, misinformation, mental health signals, polarization. This is necessary work. But designing systems primarily around harm detection produces a distorted picture of online discourse. Recent dataset work on hope speech (Sharqawi and Zaghouani, 2026; Zaghouani and Biswas, 2025), women's empowerment discourse (Zaghouani et al., 2026), and social cohesion (Ali Al-Athba and Zaghouani, 2026) reflects a deliberate shift toward also modeling constructive discourse. You cannot design interventions to support something you have not characterized, and a research literature on harm without a corresponding literature on flourishing is an incomplete basis for policy.

**Arabic LLM evaluation as a first-class research priority.** Large language models are being deployed in Arabic-speaking contexts at scale, yet systematic evaluation infrastructure does not yet exist. PalmX (Biswas et al., 2025) and AraGenEval (Biswas et al., 2025) are early steps. Arabic LLM evaluation must at minimum span four axes: dialectal coverage across Gulf, Levantine, Egyptian, Maghrebi, and Sudanese varieties; cultural fidelity; safety auditing for harms Arabic-speaking communities would recognize but English reviewers would not; and factual accuracy on Arab history and current affairs. Two decades of dialectally diverse, socially grounded Arabic corpora are exactly the substrate this agenda needs.

**Taking the translation problem seriously.** The digital citizenship project underway attempts to translate findings into something that affects the media literacy of young Arabic speakers in Qatar. Whether it succeeds will depend on pedagogical and institutional work far more than on NLP. For most of this program's history, a successful project produced publications and trained students. Those outcomes remain entirely inside academia. Building something that reaches a 14-year-old in a Doha school navigating the phenomena this program has spent a decade studying requires a different theory of impact and a willingness to be evaluated by criteria unrelated to citation counts. That shift is overdue.

## 7 Key Takeaways for the NLP Community

**A dataset without a community is an archive.** Impact comes from the organizational and community-building work surrounding a release, not from the data alone.

**Run the shared task before your models are ready.** Tasks designed before the field has converged on a problem force honest engagement with ambiguity that post-hoc evaluation cannot surface.

**Prestige-variety infrastructure does not transfer downward automatically.** The formal standard of a language is not the language. Building NLP for the prestige variety first and adapting later consistently underserves the communities who most need the technology.

**The hardest competencies for socially oriented NLP are the ones NLP programs do not teach.** Clinical partnership, ethical governance, and policy translation determine whether technically sound work produces any effect outside a benchmark. Integrating these competencies from the start of a project, not after the technical work is done, is the practical change that follows from twenty years of this experience.

## 8 Conclusion

What went wrong: assuming infrastructure automatically produces impact; treating annotation disagreement as noise; skipping clinical governance in mental health work; chasing shared task breadth over depth; and treating MSA-to-dialect transfer as a minor engineering task. What was surprising: how much community building mattered relative to any individual resource, and how consistently the hardest obstacles were social rather than technical. What comes next: modeling flourishing alongside harm, rigorous Arabic LLM evaluation, and translating findings into tools that reach communities outside academia.

None of these conclusions are uniquely Arabic. Any researcher building NLP for an underserved language community will recognize the compressed timelines, the annotation philosophy that travels from prestige registers to vernacular ones without sufficient scrutiny, the gap between detection and deployment, and the community-building work that turns individual resources into durable fields. The Arabic experience does not solve those problems. It has accumulated enough honest failure, and enough honest success, to be worth learning from.

## Limitations

This paper reflects on a twenty-year research program through the perspective of one researcher and therefore necessarily represents a partial account. Large collaborative programs evolve through the contributions of many students, collaborators, and institutional partners. Those collaborators would likely emphasize different episodes, successes, or failures, and may interpret some of the lessons differently. The narrative presented here should therefore be read as a reflective synthesis rather than a definitive historical record of the program.

A second limitation concerns the evidentiary basis of the claims. The arguments in this paper are grounded primarily in accumulated experience rather than systematic empirical analysis across multiple programs or language communities. While many of the lessons described here likely generalize to other low-resource language contexts, this generalization has not been formally tested. Researchers working on African languages, South Asian languages, or Indigenous language communities may encounter similar structural challenges but also face distinct institutional and sociolinguistic conditions that shape infrastructure development differently.

Third, the paper focuses heavily on research infrastructure, dataset development, and community organization within the academic NLP ecosystem. It does not provide a systematic evaluation of downstream real-world impact. As discussed in Section 4.1, the gap between research outputs and deployment remains large, and the extent to which any particular dataset or model influences policy, education, or public discourse is difficult to measure. This paper therefore evaluates impact primarily through indicators internal to the research community (shared tasks, collaborations, student training, and dataset reuse) rather than through external social outcomes.

A fourth limitation concerns retrospective interpretation. Reflective papers inevitably introduce narrative coherence into events that were experienced as messy and contingent when they occurred. Some failures appear obvious in hindsight but were not obvious at the time, given the constraints under which projects were conducted. Similarly, some successes described here may partly reflect broader shifts in the field rather than decisions made within this specific research program.

Finally, the analysis emphasizes structural and institutional lessons rather than detailed technical analysis. Readers seeking technical evaluation of specific models, datasets, or annotation frameworks should consult the original publications referenced throughout the paper. The purpose of this article is not to document individual technical contributions but to extract higher-level insights about the process of building NLP infrastructure and research communities over an extended period.

These limitations should be understood not as weaknesses but as boundaries on the type of claim the paper is making. The goal is to contribute a practice-based perspective that complements more formal analyses of dataset design, evaluation methodology, and NLP system development.

## Ethical Considerations

Several strands of work discussed in this paper involve sensitive forms of data collection and analysis, particularly research on mental health, hate speech, online harassment, and political discourse. These areas raise ethical questions related to privacy, consent, annotator wellbeing, and potential misuse of computational tools.

### 8.1 Privacy and Data Governance

Much of the research described here relies on publicly available social media data. Although such data are technically accessible, their use still raises important ethical questions regarding user expectations of privacy and the potential for unintended harm. Individuals posting online may not anticipate that their content will be aggregated into research datasets or analyzed by automated systems.

To mitigate these concerns, responsible data practices should include data minimization, removal of personally identifiable information when possible, and careful documentation of collection protocols. When datasets are released publicly, documentation such as datasheets or data statements should clearly describe the data sources, collection procedures, and known limitations. These practices help ensure transparency and allow downstream users to understand the context in which the data were created.

### 8.2 Mental Health Research and Risk of Harm

Research on depression detection and other mental health signals from social media text raises particularly serious ethical concerns. Predictive systems that attempt to infer psychological states from language can generate false positives or false negatives with significant consequences if deployed in real-world contexts.

As discussed in Section 4.1, a key lesson from early work in this area is that technical model development alone is insufficient. Responsible research in this domain requires collaboration with clinical experts, ethical review structures, and clear protocols for how predictions would be used in practice. Without these safeguards, there is a risk that models could be misinterpreted as diagnostic tools or used in ways that harm the individuals they aim to help.

### 8.3 Annotator Wellbeing

Annotation work involving hate speech, harassment, or traumatic content can impose psychological burdens on annotators. Exposure to harmful language or disturbing material over extended periods can negatively affect mental health. Ethical dataset construction therefore requires attention not only to annotation quality but also to the wellbeing of the people performing the annotation.

Practical safeguards may include limiting daily exposure to harmful content, providing clear opt-out mechanisms for annotators, offering mental health resources, and designing annotation workflows that distribute difficult tasks across teams rather than concentrating them on a few individuals.

### 8.4 Bias, Representation, and Cultural Context

Arabic is a linguistically and culturally diverse language family encompassing dozens of dialects across multiple regions. Datasets that overrepresent certain dialects, social groups, or geopolitical contexts risk producing models that perform unevenly across communities. Similarly, annotation decisions about hate speech, offensiveness, or political framing are shaped by cultural and demographic perspectives.

Ethical NLP research should therefore treat annotation disagreement and demographic variation as important signals rather than as noise to be eliminated. Where possible, dataset documentation should describe annotator backgrounds and dataset composition so that users can interpret model behavior appropriately.

### 8.5 Dual-Use Risks

Finally, tools developed to detect harmful content or analyze public discourse may also be used for surveillance or censorship. Systems designed to identify political framing, online dissent, or controversial speech could potentially be repurposed in ways that restrict legitimate expression.

Researchers cannot fully control how computational tools are used after publication, but awareness of dual-use risks should inform dataset design, documentation, and release decisions. Clear documentation of intended research uses and limitations can help reduce the likelihood that models are interpreted as authoritative decision-making systems.

Taken together, these considerations highlight a broader point emphasized throughout this paper: the ethical challenges of socially oriented NLP research are not peripheral concerns but central design constraints. Addressing them requires interdisciplinary collaboration among NLP researchers, social scientists, clinicians, and policy experts from the earliest stages of a project.

## Acknowledgments

Some of the projects and research activities reported in this paper received partial support from the Qatar National Research Fund (QNRF) and QRDI under grants MCSC 02-0217-250013, NPRP 14C-0916-210015, NPRP 13S-0206-200281, CWSP 18-W-0206-20044, NPRP 11S-1112-170006, and NPRP 9-175-1-033.